\documentclass{article}

\usepackage{microtype}
\usepackage{graphicx}
\usepackage{subcaption}
\usepackage{booktabs} 

\usepackage{hyperref}
\usepackage{enumitem}

\usepackage{amsthm}
\newtheorem{prop}{Proposition}

\usepackage{tikz}
\usepackage{pgfplots}
\usepackage{siunitx}
 \usepackage[preprint]{icml2026}

\usepackage{amsmath}
\usepackage{amssymb}
\usepackage{mathtools}
\usepackage{amsthm}

\usepackage[capitalize,noabbrev]{cleveref}

\theoremstyle{plain}

\theoremstyle{definition}

\theoremstyle{remark}

\usepackage[textsize=tiny]{todonotes}

\icmltitlerunning{Deep Leakage with Generative Flow Matching Denoiser}

\begin{document}

\twocolumn[
  \icmltitle{Deep Leakage with Generative Flow Matching Denoiser}



  \icmlsetsymbol{equal}{*}

  \begin{icmlauthorlist}
    \icmlauthor{Isaac Baglin}{yyy}
    \icmlauthor{Xiatian Zhu}{yyy}
    \icmlauthor{Simon Hadfield}{yyy}
  \end{icmlauthorlist}

  \icmlaffiliation{yyy}{CVSSP, University of Surrey, Guildford, United Kingdom}

\icmlcorrespondingauthor{Isaac Baglin}{ib00304@surrey.ac.uk}
\icmlcorrespondingauthor{Xiatian Zhu}{xiatian.zhu@surrey.ac.uk}
\icmlcorrespondingauthor{Simon Hadfield}{s.hadfield@surrey.ac.uk}

  \icmlkeywords{Machine Learning, ICML}

  \vskip 0.3in
]



\printAffiliationsAndNotice{}  

\begin{abstract}
Federated Learning (FL) has emerged as a powerful paradigm for decentralized model training, yet it remains vulnerable to deep leakage (DL) attacks that reconstruct private client data from shared model updates. While prior DL methods have demonstrated varying levels of success, they often suffer from instability, limited fidelity, or poor robustness under realistic FL settings. We introduce a new DL attack that integrates a generative Flow Matching (FM) prior into the reconstruction process. By guiding optimization toward the distribution of realistic images (represented by a flow matching foundation model), our method enhances reconstruction fidelity without requiring knowledge of the private data. Extensive experiments on multiple datasets and target models demonstrate that our approach consistently outperforms state-of-the-art attacks across pixel-level, perceptual, and feature-based similarity metrics. Crucially, the method remains effective across different training epochs, larger client batch sizes, and under common defenses such as noise injection, clipping, and sparsification. Our findings call for the development of new defense strategies that explicitly account for adversaries equipped with powerful generative priors.
\end{abstract}

\vspace{-25pt}
\section{Introduction}
\label{sec:intro}

FL has emerged as a decentralized paradigm for training Machine Learning (ML) models across multiple clients without requiring direct access to raw data \cite{mcmahan2017communication}. Instead of pooling sensitive information on a central server, each client performs local training on its private dataset and communicates only the resulting model updates, typically gradients or parameters, to a coordinating server. The server then aggregates these updates to produce a global model that benefits from diverse data sources while ostensibly preserving user privacy.

Two widely adopted aggregation schemes in FL are FedSGD and FedAvg \cite{mcmahan2017communication}. In FedSGD, clients transmit gradient updates after every local iteration, enabling frequent synchronization but introducing significant communication overhead. By contrast, FedAvg allows clients to perform several local training steps before sharing updated model weights, thereby reducing communication costs. However, this comes at the expense of potential model staleness due to delayed synchronization \cite{rodio2024fedstaleleveragingstaleclient,karimireddy2021scaffoldstochasticcontrolledaveraging}. Despite the privacy-preserving promise of FL, it remains susceptible to gradient-based privacy attacks. Notably, Deep Leakage, in which adversaries optimize dummy inputs so that their gradients align with those shared by clients, ultimately reconstructing training samples that closely resemble the originals \cite{zhu2019deep}. Such attacks can succeed under different threat models, including a man-in-the-middle adversary who intercepts communications \cite{zhao2022deepleakagemodelfederated} and an honest-but-curious (HbC) server that faithfully executes the FL protocol but inspects updates to extract sensitive information \cite{geiping2020inverting}. The HbC server is assumed to have full access to client updates, knowledge of the model architecture, loss function, and training procedure. By leveraging this information, the server can solve an inverse optimization problem to reconstruct private training samples.
    \begin{figure}
        \centering
        \includegraphics[width=1.1\linewidth]{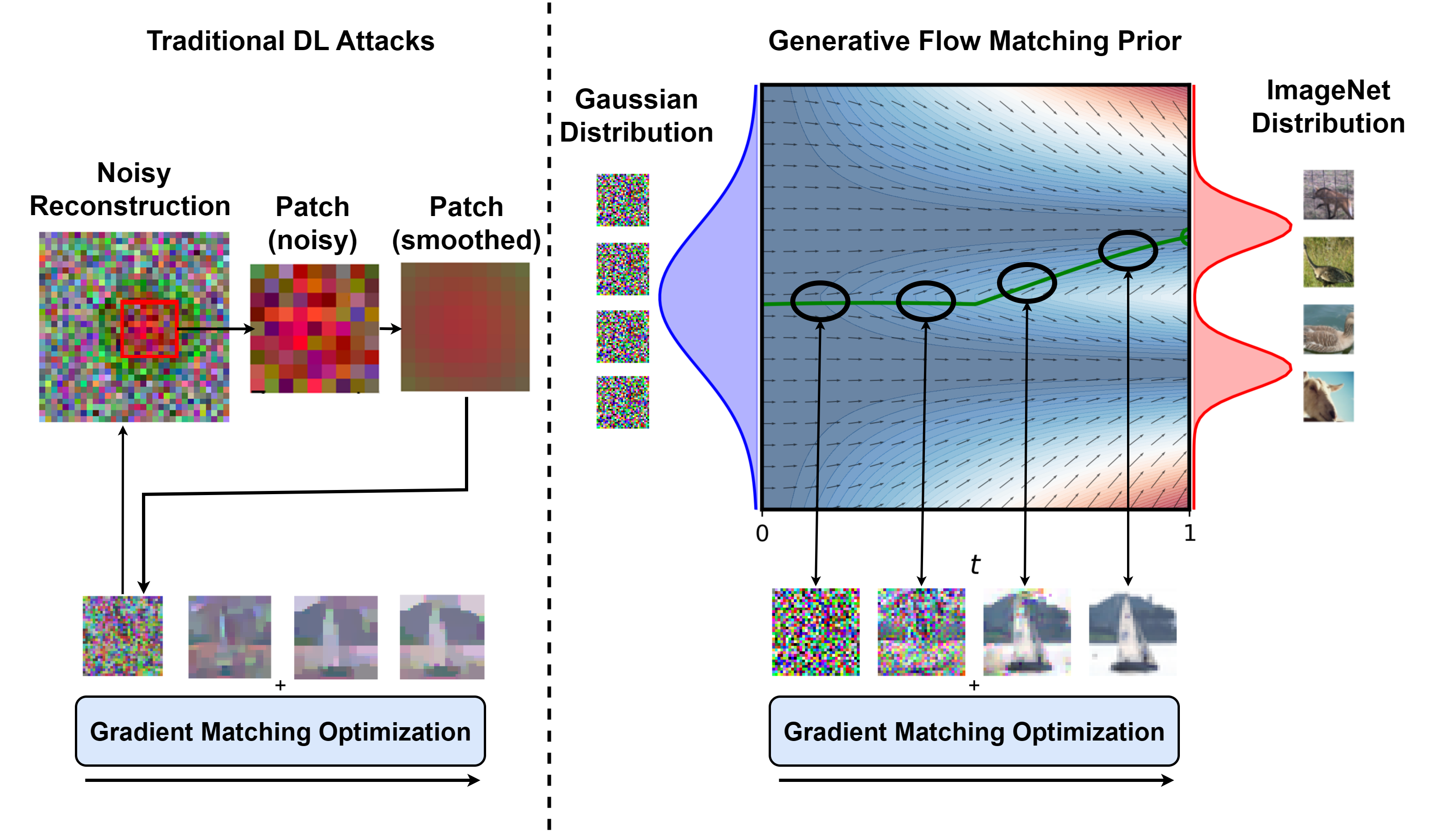}
        \caption{Left: Traditional deep leakage attacks rely on denoising, which oversmooth and remove fine details. Right: Our method leverages a flow-matching prior to denoise while preserving realism and guiding reconstructions toward natural image distributions.}
        \label{fig:placeholder}  \vspace{-10pt}
    \end{figure}
While existing deep leakage attacks have made important progress, they still face a fundamental trade-off between realism and faithfulness. Classical inversion methods that rely heavily on traditional denoising tend to oversmooth the reconstruction, suppressing high-frequency details and producing blurry images that lack semantic clarity \cite{Sun_2021}. In contrast, recent approaches that incorporate generative priors such as GAN or diffusion-based models often succeed in producing visually realistic samples \cite{jeon2021gradientinversiongenerativeimage,li2024exploringuserlevelgradientinversion}. However, these reconstructions frequently drift away from the actual training data, yielding images that are class-consistent but not faithful to the original private sample. This deviation undermines the privacy risk, since the adversary recovers “plausible” images rather than the true sensitive information.

Our method directly addresses both of these limitations by reformulating a flow matching foundation model as a learned denoiser, and integrating it into the inversion process \cite{lipman2023flowmatchinggenerativemodeling}. Unlike conventional denoising, which indiscriminately smooths out detail, our approach regularizes reconstructions toward the natural image distribution while still anchoring them closely to the gradient information contained in model updates \cite{Sun_2021,geiping2020inverting}. This dual guidance enables reconstructions that are both realistic and faithful, preserving fine-grained structure without drifting into unrelated but visually plausible modes. Our key contributions are as follows: 
\vspace{-3pt}
\begin{itemize}[topsep=2pt, itemsep=2pt, parsep=0pt]
    \item \textbf{Generative flow matching prior for leakage attacks.} We are the first to incorporate a generative flow matching prior into DL attacks. While previous methods have relied on GANs, diffusion models, or classical denoising priors \cite{Sun_2021,jeon2021gradientinversiongenerativeimage}, our FM approach uniquely models vector-field trajectories to guide reconstruction, resulting in both realistic and faithful reconstructions, a direction that, to the best of our knowledge, has not been explored in the context of DL privacy attacks.

    \item \textbf{Cross-distribution applicability.} Unlike prior generative approaches, which assume access to sufficient data from the private distribution to train high-quality diffusion or GAN models, our method leverages a flow matching foundation model trained on an unrelated dataset\cite{li2024exploringuserlevelgradientinversion,li2022auditingprivacydefensesfederated}. This relaxes attacker assumptions while still enabling high-fidelity reconstructions, significantly broadening the practicality of DL attacks.  

    \item \textbf{A reproducible evaluation framework under realistic FL settings.} We verify the viability of our attack across practical federated learning scenarios (different training epochs, models, batch sizes, State-Of-The-Art (SOTA) attacks and defenses) and provide a reproducible framework that can serve as a benchmark for future work on developing and evaluating much-needed defenses.
\end{itemize}

\section{Related Work}
\label{sec:lit}

The Deep Leakage from Gradients (DLG) attack, introduced by Zhu et al., was the first approach to reconstruct images from gradients \cite{zhu2019deep}. In this attack, a dummy image is iteratively optimized to minimize the Euclidean distance between its gradients and the real gradients available to an attacker. By leveraging L-BFGS to refine the dummy image, DLG was able to recreate images from a variety of datasets but was limited to working with gradients from randomly initialized networks \cite{cifar,LFWTech}. This reliance on untrained networks restricted its practical applications.

Geiping et al. introduced the Inverting Gradients approach, which incorporated total variation (TV) regularization to denoise reconstructed images \cite{geiping2020inverting}. This advancement enabled successful attacks on trained networks, which typically present gradients with smaller magnitudes that complicate reconstruction. While TV improves stability, it also tends to oversmooth reconstructions, suppressing high-frequency details and leading to blurry results \cite{Sun_2021}. Building upon these ideas, Yin et al. developed DeepInversion, implementing more sophisticated regularization methods by combining $\ell_{2}$ norm regularization with a novel batch normalization (BN) regularization term, which utilized the mean and variance of feature maps for improved recovery \cite{yin2020dreaming}. In their subsequent work on GradInversion, they integrated these regularization techniques with a unique group consistency term, which optimized multiple random seeds concurrently to create a consensus image \cite{yin2021see}. This method allowed attackers to effectively penalize outliers during reconstruction and proved highly effective with larger batch sizes and more complex datasets, such as ImageNet \cite{deng2009imagenet}.

 Dimitrov et al. designed Data Leakage in Federated Averaging (DLF) specifically for the FedAvg protocol \cite{dimitrov2022data}. By simulating client-side training steps and introducing an epoch-order invariant prior, DLF improves reconstruction stability across different synchronization schedules. However, this simulation-based approach struggles to recover fine grained details at later training states. Building from DLF, Zhu et al. recently proposed Surrogate Model Extension (SME), which interpolates between pre and post update weights to form “surrogate states” that approximate hidden intermediate models \cite{zhu2023surrogatemodelextensionsme}. 

Li et al. introduced Generative Gradient Leakage (GGL), which incorporates GAN-based priors into DL attacks \cite{li2022auditingprivacydefensesfederated}. By constraining reconstructions to the distribution of images learned by a GAN, GGL produces visually coherent samples. However, much like other generative approaches \cite{jeon2021gradientinversiongenerativeimage,li2024exploringuserlevelgradientinversion}, the GAN is trained on the same dataset as the victim, requiring the attacker to already possess sufficient data from the private distribution. This assumption undermines one of the central motivations for gradient leakage attacks namely, reconstructing sensitive client data to construct a dataset that would otherwise be inaccessible. Moreover, while the GAN prior improves realism, it often sacrifices fidelity, producing class-consistent but inaccurate reconstructions of the true private samples.

Our approach directly addresses these limitations by incorporating a generative FM prior that both regularizes reconstructions toward natural image distributions and preserves the fine-grained details encoded in gradients. This enables reconstructions that are simultaneously realistic, faithful, and robust across practical FL conditions.

While Deep Leakage attacks have demonstrated the vulnerability of Federated systems to privacy breaches, a range of defensive strategies have been developed to counteract these risks, aiming to enhance data security while maintaining model accuracy. Early works explore using Differential Privacy (DP) methods to protect systems against Deep Leakage, including adding noise to gradients, pruning gradients, increasing batch sizes, and using higher-resolution images \cite{zhu2019deep}. Fang et al. introduced PFMLP, a strategy that encrypts gradients with homomorphic encryption before they are transmitted \cite{fang2021privacy}. Although this approach limits accuracy by only 1\%, it significantly increases computational demands, prompting the authors to employ an optimized Paillier algorithm. Zheng et al. demonstrated that adding dropout layers before the classifier could effectively prevent Deep Leakage, however, Scheliga et al. introduced the Dropout Inversion Attack (DIA) \cite{scheliga2023dropout}. This attack jointly optimizes dropout masks, initialized from a random distribution, along with dummy data to closely mimic the client’s private example. DIA sometimes outperformed cases where dropout was absent. Scheliga et al. also introduced PRECODE, a model extension that shields gradients from Deep Leakage by concealing the latent feature space through the use of a variational bottleneck \cite{scheliga2022precode}.
\vspace{-10pt}
\section{Methodology}
\label{method}

In this section, we describe our approach for reconstructing private client data from weight updates in FL (see Figure \ref{fig:placeholder}). Our method integrates generative flow models as a learned denoiser with interpolation-based gradient matching to improve the fidelity of deep leakage attacks. We first review the generative flow matching framework and how it can be leveraged as a strong denoiser. Next, we present an interpolation-based strategy for approximating gradients from aggregated weight updates, addressing the limitations of naive gradient inversion in the federated averaging setting. Finally, we unify the learned denoising flow field with the iterative refinement of the deep-leakage attack leading to high-quality reconstructions of sensitive client data.

\subsection{Generative Flow Matching}

Flow matching is a framework for generative modeling that formulates the generation process as a continuous time flow \cite{lipman2023flowmatchinggenerativemodeling}. Let $\mathbf{x}_t \in \mathbb{R}^d$ denote the state of a sample at time $t \in [0,1]$, where $\mathbf{x}_0 \sim p_0$ is a simple base distribution (e.g., Gaussian) and $\mathbf{x}_1 \sim p_1$ is a target data distribution. The goal is to learn vector field $v_\theta(\mathbf{x}, t)$ such that trajectories of the ordinary differential equation (ODE)
\vspace{-5pt}
\begin{equation}
\frac{d \mathbf{x}_t}{d t} = v_\theta(\mathbf{x}_t, t), \quad \mathbf{x}_0 \sim p_0
\label{eq:ode}
\end{equation}
\vspace{-5pt}
transport noise samples to realistic data samples.
\vspace{-5pt}
\subsubsection{Flow Matching Objective}
\vspace{-5pt}
Flow matching minimizes the discrepancy between the learned vector field $v_\theta$ and the instantaneous velocity of a reference flow $u(\mathbf{x}_t, t)$ along trajectories from $\mathbf{x}_0$ to $\mathbf{x}_1$. A common choice is linear interpolation:
\begin{equation}
\mathbf{x}_t = (1-t)\mathbf{x}_0 + t \mathbf{x}_1, \quad
\frac{d \mathbf{x}_t}{d t} =u(\mathbf{x}_t, t) = \mathbf{x}_1 - \mathbf{x}_0.
\end{equation}
This formulation has attractive properties compared to diffusion based denoisers. The linear interpolation paths encourage direct flows which can greatly reduce the number of denoising steps required. The training loss is then
\begin{equation}
\mathcal{L}(\theta) = \mathbb{E}_{t \sim \mathcal{U}[0,1], \mathbf{x}_t \sim p_t}
\left[ \left\| v_\theta(\mathbf{x}_t, t) - u(\mathbf{x}_t, t) \right\|_2^2 \right],
\label{eq:flow_loss}
\end{equation}
where $p_t$ is the marginal distribution along the trajectories. Minimizing this loss encourages $v_\theta$ to produce flows consistent with the reference, effectively pushing noise samples toward the data manifold. Flow matching models are typically used for sampling generative examples by solving Equation \eqref{eq:ode} from noise to data, but in our work we reformulate it as a denoiser for Deep Leakage attacks.
\vspace{-5pt}
\subsection{Interpolation-Based Gradient Matching}
Deep Leakage poses a serious threat to Federated Learning, as it allows an adversary to reconstruct clients' private data from shared model updates. We consider the standard Federated Averaging (FedAvg) framework \cite{mcmahan2017communication}, where a global model with parameters $w \in \mathbb{R}^p$ is distributed to a set of clients. Each client $i$ holds a private dataset
\begin{equation}
D_i = \{(x_j, y_j)\}_{j=1}^{N_i}.
\label{eq:client_dataset}
\end{equation}
and performs local training for $E$ epochs using mini-batch stochastic gradient descent (SGD) with batch size $B$. After $T_i = E \cdot \lceil N_i / B \rceil$ local steps, the client transmits the updated weights $w_T^i$ and the dataset size $N_i$ back to the server. 

From the adversary’s perspective, the observable information for a given client is the initial weights $w_0$ (received from the server), the locally updated weights $w_T$, and the number of local samples $N$. The corresponding weight update is
\begin{equation}
\Delta w = w_T - w_0.
\end{equation}
\vspace{-25pt}
\subsubsection{Limitations of Naive Gradient Inversion}
\vspace{-5pt}
Classic gradient inversion attacks, such as Deep Leakage from Gradients (DLG)~\cite{zhu2019deep} and its improvements, assume access to the exact gradient
\begin{equation}
 \frac{1}{N} \sum_{j=1}^N \nabla_{w_0} \ell(w_0, x_j).
\label{eq:true_grad}
\end{equation}
where $\nabla_{w_0}$ denotes the gradient with respect to the training loss $\ell(w_0, x_j)$ evaluated on sample $x_j$.  
However, in the FedAvg setting, the observed weight update $\Delta w$ corresponds to an \emph{accumulation} of many gradients along the trajectory $\{w_t\}_{t=0}^T$. Consequently, treating $\Delta w$ as a proxy gradient at $w_0$ introduces a bias that increases with the number of local steps $T$, leading to degraded reconstruction quality~\cite{dimitrov2022data}. 
\vspace{-5pt}
\subsubsection{Surrogate Matching Loss}
\vspace{-5pt}
Building on the rationale of the Surrogate Model Extension (SME) \cite{zhu2023surrogatemodelextensionsme} and Deep Leakage from Federated Averaging (DLF) \cite{dimitrov2022data}, we adopt an interpolation-based reconstruction strategy. The key idea, is to introduce a \emph{surrogate model} that lies between the initial and final weights. Formally, we parameterize the surrogate as
\begin{equation}
\hat{w} = \alpha w_0 + (1-\alpha) w_T, \qquad \alpha \in [0,1].
\end{equation}
At this surrogate model, we generate dummy data $\hat{x}$ and compute its gradient $\nabla_{\hat{w}} \ell(\hat{w}, \hat{x})$. To enforce consistency with the observed update $\Delta w$, we minimize the cosine similarity loss
\begin{equation}
\mathcal{L}_{\text{sim}}(\hat{x}, \hat{w}) 
= 1 - \frac{\langle -\Delta w, \nabla_{\hat{w}} \ell(\hat{w}, \hat{x}) \rangle}{\|\Delta w\| \cdot \|\nabla_{\hat{w}} \ell(\hat{w}, \hat{x})\|}.
\end{equation}
The central intuition is that although $\Delta w$ is not itself a gradient, there exists (or approximately exists) a point $\hat{w}$ along the interpolation between $w_0$ and $w_T$ such that the gradient of the true data $D$ at $\hat{w}$ is nearly parallel to $\Delta w$ \cite{zhu2023surrogatemodelextensionsme}. 
\vspace{-5pt}
\subsection{Flow Matching Regularization Loss}
\vspace{-5pt}
A central challenge in gradient-based data reconstruction is that standard
regularizers such as total variation (TV) or $\ell_2$ penalties, while effective for stabilizing optimization, often lead to overly smooth and visually unrealistic images \cite{Sun_2021}. For example, TV minimization encourages piecewise constant reconstructions, which suppresses high-frequency details such as textures and edges. Similarly, perceptual losses based on feature distances in pretrained vision networks can mitigate oversmoothing but still tend to bias reconstructions toward generic solutions. In adversarial reconstruction settings, these deficiencies manifest as \emph{washed-out faces}, \emph{blurred digits}, or \emph{checkerboard artifacts}. Thus, a stronger prior is required to guide optimization toward the true data manifold.

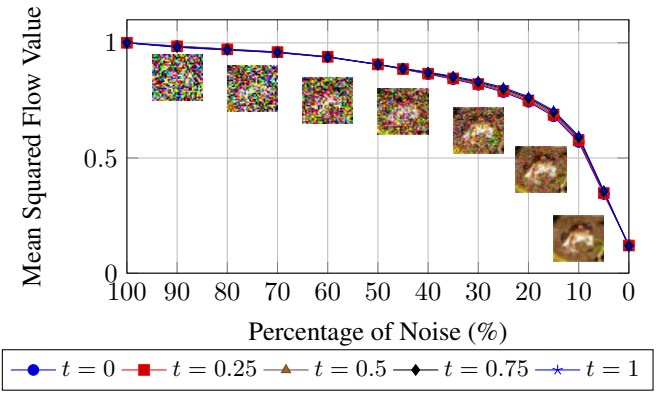
\begin{figure}[h!]
\centering
\begin{tikzpicture}
\begin{axis}[
    width=\linewidth,
    height=0.6\linewidth,
    xlabel={Percentage of Noise (\%)},
    ylabel={Mean Squared Flow Value},
    xmin=0, xmax=100,
    x dir=reverse,
    ymin=0, 
    xtick={0,10,20,30,40,50,60,70,80,90,100},
    legend style={
    at={(0.4,-0.3)},
    anchor=north,
    legend columns=5,
    legend cell align=left,
    column sep=0.05pt,
    font=\small,
},
    grid=both,
]


\addplot+[mark=*] coordinates {
    (0,0.1215)
    (5,0.3436)
    (10,0.5696)
    (15,0.6811)
    (20,0.7440)
    (25,0.7861)
    (30,0.8177)
    (35,0.8422)
    (40,0.8642)
    (45,0.8856)
    (50,0.9061)
    (60,0.9397)
    (70,0.9602)
    (80,0.9728)
    (90,0.9861)
    (100,1.00)
};
\addlegendentry{$t=0$}

\addplot+[mark=square*] coordinates {
    (0,0.1202)
    (5,0.3480)
    (10,0.5767)
    (15,0.6875)
    (20,0.7496)
    (25,0.7910)
    (30,0.8212)
    (35,0.8457)
    (40,0.8660)
    (45,0.8863)
    (50,0.9063)
    (60,0.9393)
    (70,0.9596)
    (80,0.9718)
    (90,0.9848)
    (100,1)
};
\addlegendentry{$t=0.25$}

\addplot+[mark=triangle*] coordinates {
    (0,0.1189)
    (5,0.3521)
    (10,0.5838)
    (15,0.6941)
    (20,0.7554)
    (25,0.7961)
    (30,0.8256)
    (35,0.8484)
    (40,0.8681)
    (45,0.8872)
    (50,0.9063)
    (60,0.9388)
    (70,0.9589)
    (80,0.9708)
    (90,0.9836)
    (100,1)
};
\addlegendentry{$t=0.5$}

\addplot+[mark=diamond*] coordinates {
    (0,0.1177)
    (5,0.3553)
    (10,0.5901)
    (15,0.7003)
    (20,0.7609)
    (25,0.8011)
    (30,0.8299)
    (35,0.8519)
    (40,0.8702)
    (45,0.8881)
    (50,0.9064)
    (60,0.9383)
    (70,0.9582)
    (80,0.9698)
    (90,0.9821)
    (100,1)
};
\addlegendentry{$t=0.75$}

\addplot+[mark=star] coordinates {
    (0,0.1164)
    (5,0.357)
    (10,0.5953)
    (15,0.706)
    (20,0.7659)
    (25,0.806)
    (30,0.8338)
    (35,0.8548)
    (40,0.8722)
    (45,0.889)
    (50,0.9064)
    (60,0.9378)
    (70,0.9576)
    (80,0.9689)
    (90,0.9806)
    (100,1)
};
\addlegendentry{$t=1$}

\addplot graphics[
    xmin=5, xmax=15,
    ymin=0.05, ymax=0.25
] {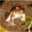};

\addplot graphics[
    xmin=12.5, xmax=22.5,
    ymin=0.35, ymax=0.55
] {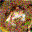};

\addplot graphics[
    xmin=25, xmax=35,
    ymin=0.52, ymax=0.72
] {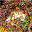};

\addplot graphics[
    xmin=40, xmax=50,
    ymin=0.6, ymax=0.8
] {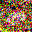};

\addplot graphics[
    xmin=55, xmax=65,
    ymin=0.65, ymax=0.85
] {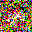};

\addplot graphics[
    xmin=70, xmax=80,
    ymin=0.7, ymax=0.9
] {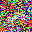};

\addplot graphics[
    xmin=85, xmax=95,
    ymin=0.75, ymax=0.95
] {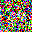};

\end{axis}
\end{tikzpicture}
\caption{Effect of noise on the Mean Squared Flow (MSF) of a Flow Matching model. As the noise level decreases, the MSF correspondingly decreases, indicating that Flow Matching can serve as an effective denoising prior.}
\label{flow_proof}
\end{figure}

Recent advances in flow matching generative models suggest that their learned vector fields $v_\theta(\mathbf{x},t)$ capture rich structural information about natural images \cite{lipman2023flowmatchinggenerativemodeling}. Intuitively, when $\mathbf{x}$ lies close to the data manifold, the vector field is small and smoothly varying, since only mild adjustments are needed to align $\mathbf{x}$ with the plausible data manifold. In contrast, unrealistic or noisy images typically induce large, erratic velocities, reflecting the model's attempt to transport them back to the manifold (Figure \ref{flow_proof}). This observation motivates the following heuristic: \emph{the magnitude of the flow field is correlated with the distance of a sample from the data distribution.}
\vspace{-20pt}
\subsubsection{Assumptions}
We formalize this intuition with two assumptions:
\vspace{-10pt}

\begin{enumerate}[label=(A\arabic*)]
    \item \textbf{Flow regularity (Lipschitz continuity).}  
    For any two images $\mathbf{x}_1, \mathbf{x}_2 \in \mathbb{R}^d$ and any $t \in [0,1]$,
    \begin{equation}
\|v_\theta(\mathbf{x}_1,t) - v_\theta(\mathbf{x}_2,t)\|_2
\;\leq\; L \|\mathbf{x}_1 - \mathbf{x}_2\|_2
\end{equation}
    for some constant $L > 0$.  
    This inequality states that if two inputs are visually similar, their predicted flow vectors are also similar, up to a linear factor. In other words, the vector field does not change erratically when the input changes only slightly.

\item \textbf{Velocity–distance alignment.}  
Let $\mathcal{M}$ denote the data manifold of natural images, and let $d(x) = \mathrm{dist}(x,\mathcal{M})$ be the distance of $x$ to the closest point on $\mathcal{M}$. We assume that the predicted velocity field is approximately aligned with the manifold normal ($n(x)$) and its magnitude correlates with the distance:
\begin{equation}
v_\theta(x,t) \;\approx\; -\,\psi\bigl(d(x)\bigr)\,n(x),
\end{equation}
where $\psi : \mathbb{R}_{\ge 0} \to \mathbb{R}_{\ge 0}$ is a non-decreasing function. Intuitively, this means that images close to the data distribution induce only small corrective flows, while images farther away are pushed back more strongly toward $\mathcal{M}$, in the direction normal to the manifold.
\end{enumerate}
\noindent
Under assumptions A1 and A2, minimizing the expected squared norm of the
velocity field acts as a means for projecting reconstructions toward the
data manifold.
\begin{prop}
Let $\hat{x} \in \mathbb{R}^d$ be a reconstruction variable, and consider the
regularizer
\begin{equation}
\mathcal{L}_{\mathrm{flow}}(\hat{x},t) = \|v_\theta(\hat{x},t)\|_2^2.
\label{eq:flow_norm}
\end{equation}
Then, for any fixed $t$, gradient descent on $\hat{x}$ with respect to $\mathcal{L}_{\mathrm{flow}}$ decreases $\mathrm{dist}(\hat{x}, \mathcal{M})$ in expectation, up to higher-order terms controlled by the Lipschitz constant $L$.
\end{prop}
\vspace{-20pt}
\begin{proof}
By assumption (A2), we have
\begin{equation}
v_\theta(\hat{x},t) \;\approx\; -\psi\bigl(d(\hat{x})\bigr)\,n(\hat{x}),
\quad d(\hat{x}) = \mathrm{dist}(\hat{x}, \mathcal{M}),
\end{equation}
where $n(\hat{x}) = \nabla_{\hat{x}} d(\hat{x})$ is the outward unit normal (i.e.~the direction from the closest point on $\mathcal{M}$ to $\hat{x}$, well-defined for $\hat{x}$ in a tubular neighborhood of $\mathcal{M}$). Substituting this expression into the definition of $\mathcal{L}_{\mathrm{flow}}$, we obtain
\begin{equation}
\small
\mathcal{L}_{\mathrm{flow}}(\hat{x},t) = \bigl\|-\psi(d(\hat{x}))\,n(\hat{x})\bigr\|_2^2 = \psi^2(d(\hat{x}))\,\|n(\hat{x})\|_2^2.
\end{equation}
Since $n(\hat{x})$ is a unit vector, $\|n(\hat{x})\|_2^2=1$, and therefore
\begin{equation}
\mathcal{L}_{\mathrm{flow}}(\hat{x},t) \;\approx\; \psi^2(d(\hat{x})).
\end{equation}
Employing the chain and product rules we can differentiate this with respect to $\hat{x}$ to give
\begin{equation}
\nabla_{\hat{x}} \mathcal{L}_{\mathrm{flow}} 
\;\approx\; 2\psi(d(\hat{x}))\,\psi'(d(\hat{x}))\,\nabla_{\hat{x}} d(\hat{x}).
\end{equation}
Since $\nabla_{\hat{x}} d(\hat{x}) = n(\hat{x})$, we can equivalently write
\begin{equation} 
\nabla_{\hat{x}} \mathcal{L}_{\mathrm{flow}} 
\;\approx\; \psi(d(\hat{x}))\,\psi'(d(\hat{x}))\,n(\hat{x}).
\end{equation}
after rescaling $\psi$ to absorb the constant factor. Thus, gradient descent on $\hat{x}$ moves it approximately along the inward normal direction $-n(\hat{x})$, thereby reducing its distance to $\mathcal{M}$. Lipschitz continuity (A1) ensures the approximation error is bounded, preventing divergence.
\end{proof}
\vspace{-10pt}
\subsubsection{Regularization objective}
We therefore define our flow-based prior as
\begin{equation}
\mathcal{L}_{\mathrm{flow}}(\hat{x}, i) = \mathbb{E}_{t_i}\left[
    \| v_\theta(\hat{x}, t)\|_2^2 \right],
\label{eq:flow_reg_new}
\end{equation}
where $t_i=i/K$ maps the optimization iteration $i \in \{0,\ldots,K\}$ into
pseudo-time. Minimizing this loss encourages the dummy image $\hat{x}$ to lie in regions of low flow magnitude, i.e.~near the data manifold captured by the pretrained model.

The overall optimization for deep leakage then combines a reconstruction loss $\mathcal{L}_{\text{sim}}$ with the flow regularization:
\begin{equation}
\small
\mathcal{L}_{\text{total}}(\hat{x}, \hat{w}, i) = \mathcal{L}_{\text{sim}}(\hat{x}, \hat{w}) + \lambda \, \mathcal{L}_{\text{flow}}(\hat{x}, i) + \sigma \, TV(\hat{x}),
\end{equation}
where $\lambda$ and $\sigma$ control the relative strength of the flow regularization and TV, respectively. Because $\mathcal{L}_{\text{flow}}$ is inherently denoising, $\sigma$ should be kept minimal. While $TV$ still offers some benefit, it no longer needs to be highly influential, as excessive weighting could risk oversmoothing the reconstruction. We perform joint optimization over the dummy image $\hat{x}$ and the interpolation parameter $\alpha$ using the Adam optimizer. The final reconstructed private data is obtained as
\vspace{-5pt}
\begin{equation} \label{eq:optim_split}
\hat{x}^* = \arg \min_{\hat{x}, \alpha} \mathcal{L}_{\text{total}},
\end{equation}
where $\hat{x}^*$ represents the recovered client data. By jointly optimizing both the image and the interpolation coefficient, and by incorporating flow regularization, the method produces high-fidelity reconstructions that are consistent with the observed weight updates.

\vspace{-10pt}
\section{Experiments}
\label{eval}
\vspace{-3pt}
The effectiveness of the proposed method in reconstructing private data from federated learning updates is systematically assessed, providing a comprehensive framework for comparing attack performance under realistic FL scenarios.

\textbf{FL Setup.} We evaluate our method on the standard federated learning classification tasks using two benchmark datasets: (1) CIFAR-10, a 10-class image classification dataset with images of size 32 × 32, and (2) Tiny-ImageNet, a 200-class image classification dataset with images of size 64 × 64 \cite{cifar,Le2015TinyIV}. For consistency with prior work, in both settings, federated clients jointly train a custom ConvNet (following the architecture used in \cite{zhu2023surrogatemodelextensionsme}). The FL framework adopts the standard FedAvg configuration, where clients perform local training (5 batches) and communicate model updates to a central server. We assume an honest-but-curious server that correctly orchestrates federated training but, in its curiosity, seeks to invert client model updates to reconstruct private data. Our evaluation considers client FL model weights at various global FL Epochs, E (from random initialization to model convergence) and a range of client batch sizes, B, in order to assess the robustness and effectiveness of the proposed attack. Note, FL training is performed using the Adam optimizer with a learning rate of 0.1 and cross-entropy loss. The flow matching model in our attack is implemented with a UNet backbone (input resolution 32×32, base channel width 64, two residual blocks per stage) \cite{lipman2023flowmatchinggenerativemodeling}. It is trained in an unconditional setting (i.e., without class conditioning), using ImageNet pretraining while the federated learning task is evaluated on CIFAR and Tiny-ImageNet \cite{deng2009imagenet}. All experiments were conducted on a workstation equipped with an NVIDIA RTX 3090 GPU (CUDA 12.9), running Ubuntu 20.04. The framework was implemented in PyTorch 2.1.

\textbf{Evaluation Metrics.} In addition to qualitative visual comparisons, we report quantitative metrics to evaluate the similarity between the target and reconstructed images. Specifically, we use: (1) \textit{Peak Signal-to-Noise Ratio} (PSNR ↑): measures the ratio between maximum pixel intensity and the mean squared reconstruction error; (2) \textit{Structural Similarity Index} (SSIM ↑): assesses structural consistency between the target and reconstructed images; (3)\textit{ Learned Perceptual Image Patch Similarity} (LPIPS ↓): computed with both VGG and AlexNet backbones, this evaluates perceptual similarity in feature space; (4) \textit{Feature Mean Squared Error} (FMSE ↓): measures the mean squared error in representation space, capturing alignment between feature embeddings of the target and reconstruction; (5) \textit{Mean Squared Error} (MSE ↓): computes the pixel-wise error between images; (6) \textit{Total Variation} (TV ↓): quantifies smoothness of the reconstructed image; (7) \textit{VRAM Usage} (↓): measures GPU memory required for the reconstruction process; and (8) \textit{FLOPS} (↓): reports the total number of floating point operations required for reconstruction, capturing computational efficiency.
\vspace{-5pt}
\begin{table}[h!]
\centering
\caption{Effect of noise standard deviation on model accuracy and deep leakage reconstruction.}
\resizebox{0.4\textwidth}{!}{%
\begin{tabular}{c|c|c}
\hline
\textbf{Std of Noise} & \textbf{Model Accuracy (\%)} & \textbf{Reconstruction SSIM} \\
\hline
0      & 70.32 & 0.621 \\
$1 \times 10^{-4}$ & 68.43 & 0.618 \\
$1 \times 10^{-3}$ & 59.77 & 0.469 \\
$2 \times 10^{-3}$ & 10.00 & 0.306 \\
$1 \times 10^{-2}$ & 10.00 & 0.070 \\
\hline
\end{tabular}
}
\vspace{-10pt}
\label{tab:noise_vs_ssim}
\end{table}

\textbf{Defense Implementations.} The defense parameters are selected to reflect the practical requirement of preserving model utility, rather than providing trivial but unusable protection. For instance, while injecting extremely large noise could in principle guarantee privacy, such settings would render the model effectively useless (see Table \ref{tab:noise_vs_ssim}). Instead, the defenses used in our evaluation include (1) Gaussian Noise:  sampled from $\mathcal{N}(0, \sigma^{2} I)$ where $\sigma$ = $1 \times 10^{-3}$, chosen to perturb updates without substantially degrading training accuracy; (2) Weight Clipping: model updates are clipped with an upper bound of 
0.05, limiting extreme updates while maintaining convergence; (3) Weight Sparsification: magnitude-based pruning with 5\% sparsity, which introduces communication efficiency gains while still allowing the model to train effectively.; and (4) Precode: a lightweight encoding of intermediate representations into a lower-dimensional latent space, which acts as an information bottleneck and mitigates the risk of direct gradient leakage while retaining sufficient task-relevant features for training \cite{scheliga2022precode}.

\textbf{SOTA Attack Implementations.} We compare our method against several state-of-the-art attack approaches:
\begin{itemize}[topsep=2pt, itemsep=2pt, parsep=0pt]
    \item \textbf{GradInversion:} improves stability and visual fidelity by minimizing an $\ell_2$ gradient matching loss, optimized with Adam, while additionally employing a consensus regularization loss through joint optimization of multiple random seeds \cite{yin2021see}.

    \item \textbf{Generative Gradient Leakage (GGL):} leverages GAN-based priors and latent space optimization, using $\ell_2$ loss with Adam to refine reconstructions \cite{li2022auditingprivacydefensesfederated}.

    \item \textbf{Data Leakage in Federated Averaging (DLF):} includes a two-part reconstruction objective: a simulation-based gradient matching loss ($L_{sim}$), which explicitly simulates the client’s local training steps, and an epoch-order invariant prior ($L_{inv}$), which regularizes reconstructions across epochs despite random batch order. Optimised with Adam \cite{dimitrov2022data} .

    \item \textbf{Surrogate Model Extension (SME):} SME constructs surrogate states as linear combinations of pre- and post-update weights. These states approximate hidden intermediate models. The method minimizes a cosine similarity based gradient matching loss with the Adam optimizer \cite{zhu2023surrogatemodelextensionsme}.
\end{itemize}

Each attack is executed until convergence, defined by the recovery threshold. This threshold is based on the gradient/weight matching loss, as it is the metric accessible to the attacker. Typical iteration counts for each method are as follows: GradInversion (500 iterations across 5 optimization seeds), GGL (500 iterations), DLF (10,000 iterations), SME (30,000 iterations), and our attack ranges between 20,000 and 30,000 iterations. The coefficients for the loss terms are adopted from the original papers; in our attack, we set $\lambda = 1.4x10^{-5}$ and $\sigma = 0.1$. 

\begin{table*}
\centering
\caption{Reconstruction performance on CIFAR-10 and Tiny-ImageNet across deep leakage attacks. Metrics are averaged over multiple reconstructions; best results are highlighted. Results taken at Epoch 2 of global training.}
\resizebox{0.9\textwidth}{!}{
\renewcommand{\arraystretch}{1}
\setlength{\tabcolsep}{3pt}
\begin{tabular}{cc|ccccccccc}
\hline
Dataset  & Attack Name & PSNR$\uparrow$ & SSIM$\uparrow$ & LPIPS(VGG)$\downarrow$ & LPIPS(Alex)$\downarrow$ & FMSE$\downarrow$ & MSE$\downarrow$ &  TV$\downarrow$  &  VRAM(GB)$\downarrow$ & TFLOPs $\downarrow$  \\
\hline
& GradInversion & 8.695 & 0.159 & 0.598  & 0.180 & 0.992  & 0.152 & 0.653 & 0.742 & \textbf{0.003}\\
& GGL & 11.415 & 0.164 & 0.505 & 0.114 & 2.948 & 0.092 & 0.092 & 2.574  & 5.677\\
CIFAR 10 & DLF & 12.708 & 0.171 & 0.608 & 0.239 & 1.827 & 0.068 & 0.332 & \textbf{0.654} & 3.660\\
& SME& 18.314 & 0.560 & 0.312 &\textbf{ 0.075} & 0.774 & 0.031 & 0.064 & 0.824 & 0.302\\
& Ours & \textbf{20.861} & \textbf{0.621} & \textbf{0.288} & 0.087 & \textbf{0.669} & \textbf{0.023} & \textbf{0.057} & 1.140 & 1.115 \\
\hline
& GradInversion& 8.291 & 0.077 & 0.569 & 0.138 & 2.341 & 0.159 & 0.594 & 0.746 & \textbf{0.003}\\
& GGL & 11.490 & 0.106 & 0.578 & 0.158 & 1.947 & 0.111 & 0.107 & 2.575 & 5.677\\
Tiny ImageNet & DLF  & 12.028 & 0.142 & 0.504 & 0.114 & 1.076 & 0.078 & 0.078 & \textbf{0.656} &3.660 \\
& SME & 15.187 & 0.423 & 0.359 & 0.064 & \textbf{0.331} & 0.036 & 0.123 & 0.824 &  0.302\\
& Ours & \textbf{15.501} & \textbf{0.444} & \textbf{0.333} & \textbf{0.050} & 0.401 & \textbf{0.034} & \textbf{0.115} & 1.142 & 1.115 \\
\hline
\end{tabular} }

\label{tab:deep_leakage_metrics}
\end{table*}
\vspace{-10pt}
\begin{table}[h!]
\centering
\caption{Comparison of different neural network architectures on image reconstruction quality.}
\resizebox{0.5\textwidth}{!}{
\renewcommand{\arraystretch}{1}
\setlength{\tabcolsep}{3pt}
\begin{tabular}{c|cccccc}
\hline
 Model architecture & PSNR$\uparrow$ & SSIM$\uparrow$ & LPIPS(VGG)$\downarrow$ & LPIPS(Alex)$\downarrow$ &  MSE$\downarrow$ &  TV$\downarrow$  \\
\hline
  VGG-11 \cite{simonyan2015deepconvolutionalnetworkslargescale}&16.514 &0.491  &0.406 &0.095  & 0.051  &0.084  \\
  MLP-Mixer \cite{tolstikhin2021mlpmixerallmlparchitecturevision}  &21.823 &0.739 &0.189 &0.033   &0.015   & 0.060\\
 ResNet-8 \cite{7780459}& 17.518 &0.576  &0.351 &0.086  &0.040  & 0.086  \\
ResNet-18 \cite{7780459} & 19.253 &0.642  &0.231 &0.037  & 0.022  &0.071  \\
Custom ConvNet \cite{zhu2023surrogatemodelextensionsme} & 20.861 & 0.621 & 0.288 & 0.087  & 0.023 & 0.057 \\
\hline

\end{tabular} } 

\label{tab:deep_leakage_models}
\end{table}

\begin{table}
\centering
\caption{Impact of different defenses (Gaussian noise, clipping, sparsification and Precode\cite{scheliga2022precode}) on SOTA DL attacks. Results taken at Epoch 2 of global training.}
\renewcommand{\arraystretch}{1}
\setlength{\tabcolsep}{3pt}
\resizebox{0.5\textwidth}{!}{%
\begin{tabular}{cc|ccccccc}
\hline
Defense Name  & Attack Name & PSNR$\uparrow$ & SSIM$\uparrow$ & LPIPS(V)$\downarrow$ & LPIPS(A)$\downarrow$ & FMSE$\downarrow$ & MSE$\downarrow$ &  TV$\downarrow$  \\
\hline
& GradInv& 5.995 & 0.023 & 0.685 & 0.365 & 12.277 & 0.306 & 0.662 \\
& GGL & 12.280 & 0.150  & 0.507  & 0.122 & 0.958  & 0.093  & 0.096  \\
Gaussian Noise & DLF  & 9.887 & 0.090 & 0.603 & 0.184 & 5.259  & 0.111  & 0.327\\
& SME & 16.402 & 0.443  & 0.394  & \textbf{0.092} & 0.875  & 0.032  & 0.062 \\
& Ours & \textbf{17.216} & \textbf{0.469} & \textbf{0.378}  & 0.094 & \textbf{0.833}  & \textbf{0.029}  & \textbf{0.058}   \\
\hline
& GradInv & 5.423 & 0.027  & 0.665  & 0.385 & 30.443  & 0.292  & 0.659  \\
& GGL& 11.425 & 0.164  & 0.505 & 0.114 & 2.947  & 0.091  & 0.096 \\
Clipping & DLF  & 9.508 & 0.060  & 0.624 & 0.228 & 1.386  & 0.077  & \textbf{0.011}  \\
& SME & 15.565 & 0.421  & \textbf{0.395}  & 0.097 & \textbf{0.801} & \textbf{0.045 } & 0.082  \\
& Ours & \textbf{16.976} & \textbf{0.492}  & 0.396  & \textbf{0.080} & 0.947  & \textbf{0.045}  & 0.081 \\
\hline
& GradInv & 7.934 & 0.058 & 0.670  & 0.255 & 2.009  & 0.171  & 0.721  \\
& GGL & 11.545 & 0.148  & 0.513  & 0.136 & 2.789  & 0.102  & 0.100   \\
Sparsification & DLF  & 9.485 & 0.049  & 0.625  & 0.226 & 6.370  & 0.129  & 0.410 \\
& SME & 16.527 & 0.512  & 0.317  & 0.072 & 0.729  & 0.033  & 0.073   \\
& Ours & \textbf{18.576} & \textbf{0.554}  & \textbf{0.284}  & \textbf{0.061} & \textbf{0.622}  & \textbf{0.024}  & \textbf{0.072} \\
\hline
& GradInv &6.207  & 0.016 &0.682  &0.317  &9.710 &0.249   &0.722  \\
& GGL & 11.872  &0.157   &0.509   &0.125  &  1.876 &  0.097 & 0.098    \\
Precode & DLF  & 9.434  &0.057  & 0.648  &0.204  &3.805   &0.129  &0.433  \\
& SME &14.712  &0.369  &0.431 &\textbf{0.094}  &\textbf{0.694}   &0.061  &0.106    \\
& Ours & \textbf{14.947} &  \textbf{0.385}  &\textbf{0.418}   &  0.095  &0.755   & \textbf{0.055}   &\textbf{0.102}  \\
\hline
\end{tabular}%
}
\vspace{-5pt}
\label{tab:deep_leakage_def_metrics}
\end{table}
\vspace{-4pt}
\subsection{Reconstruction Performance}
We systematically evaluate the effectiveness of our attack in recovering private client data from federated updates, comparing against established baselines under varying conditions of training progress, batch size, and defense mechanisms.

\textbf{Quantitative comparisons.} Table \ref{tab:deep_leakage_metrics} presents reconstruction performance across CIFAR-10 and Tiny-ImageNet. Our method consistently outperforms all baselines across nearly every metric. For CIFAR-10, we achieve a PSNR of 20.861 and SSIM of 0.621, surpassing the second best technique (SME) by a notable margin while also achieving the lowest LPIPS (0.288/0.087) and FMSE (0.669). Similar trends hold on Tiny-ImageNet, where our method maintains a leading SSIM of 0.444 despite the greater reconstruction difficulty associated with 200-class images. While SME remains competitive, particularly in terms of computational efficiency, our approach delivers a stronger balance of pixel-level fidelity and perceptual alignment. The slight increase in VRAM usage and FLOPS relative to SME is offset by significantly improved reconstructions, demonstrating a favorable trade-off.
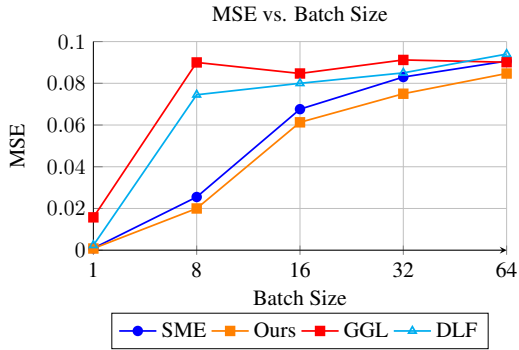
\begin{figure}[h!]
\centering
\resizebox{0.85\linewidth}{!}{%
\begin{tikzpicture}

\begin{axis}[
    width=\linewidth,
    height=0.6\linewidth,
    xlabel={Batch Size},
    ylabel={MSE},
    ymin=0, ymax=0.1,
    xtick=data,
    symbolic x coords={1,8,16,32,64},
    ytick={0,0.02,0.04,0.06,0.08,0.1},
    scaled y ticks=false, 
    tick label style={/pgf/number format/fixed}, 
    legend style={at={(0.5,-0.3)}, anchor=north, legend columns=4},
    axis y line*=left,
    axis x line=bottom,
    grid=major,
    title={MSE vs. Batch Size}
]
\addplot[blue, thick, mark=*] coordinates {
    (1,0.000927)
    (8,0.0255)
    (16,0.0676)
    (32,0.083)
    (64,0.09066)

};
\addlegendentry{SME}

\addplot[orange, thick, mark=square*] coordinates {

    (1,0.000807)
    (8,0.02)
    (16,0.0613)
    (32,0.075)
    (64,0.0847)

};
\addlegendentry{Ours}
\addplot[red, thick, mark=square*] coordinates {
    (1,0.01574)
    (8,0.09)
    (16,0.0847)
    (32,0.0912)
    (64,0.0901)

};
\addlegendentry{GGL}

\addplot[cyan, thick, mark=triangle] coordinates {
    (1,0.00236)
    (8,0.0745)
    (16,0.08)
    (32,0.085)
    (64,0.094)

};
\addlegendentry{DLF}

\end{axis}


\end{tikzpicture} }  \vspace{-5pt}
\caption{Effect of batch size on reconstruction error (MSE) at Epoch 1 of global training.}
\label{batch_fig} \vspace{-15pt}
\end{figure}
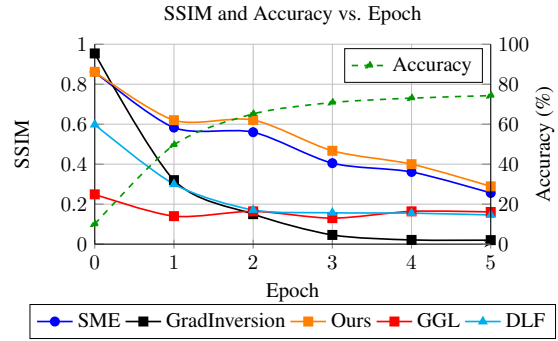
\begin{figure}
\centering
\resizebox{0.9\linewidth}{!}{%
\begin{tikzpicture}

\begin{axis}[
    width=1\linewidth,  
    height=0.6\linewidth, 
    xlabel={Epoch},
    ylabel={SSIM},
    xmin=0, xmax=5,
    ymin=0, ymax=1,
    xtick={0,1,2,3,4,5},
    ytick={0,0.2,0.4,0.6,0.8,1.0},
    legend style={at={(0.5,-0.3)}, anchor=north, legend columns=5},
    axis y line*=left,
    axis x line=bottom,
    grid=major,
    title={SSIM and Accuracy vs. Epoch},
]

\addplot[blue, thick, mark=*, smooth] coordinates {
    (0,0.862)
    (1,0.583)
    (2,0.560)
    (3,0.405)
    (4,0.361)
    (5,0.256)
};
\addlegendentry{SME}

\addplot[black, thick, mark=square*, smooth] coordinates {
    (0,0.954)
    (1,0.320)
    (2,0.15)
    (3,0.046)
    (4,0.021)
    (5,0.020)
};
\addlegendentry{GradInversion}

\addplot[orange, thick, mark=square*, smooth] coordinates {
    (0,0.862)
    (1,0.620)
    (2,0.621)
    (3,0.467)
    (4,0.400)
    (5,0.289)
};
\addlegendentry{Ours}
\addplot[red, thick, mark=square*,smooth] coordinates {
    (0,0.249)
    (1,0.140)
    (2,0.164)
    (3,0.130)
    (4,0.164)
    (5,0.161)
};
\addlegendentry{GGL}

\addplot[cyan, thick, mark=triangle*,smooth] coordinates {
    (0,0.598)
    (1,0.3)
    (2,0.171)
    (3,0.157)
    (4,0.155)
    (5,0.146)
};
\addlegendentry{DLF}

\end{axis}

\begin{axis}[
    width=1\linewidth,
    height=0.6\linewidth,
    xmin=0, xmax=5,
    ymin=0, ymax=100,
    xtick=\empty,
    ytick={0,20,40,60,80,100},
    ylabel={Accuracy (\%)},
    axis y line*=right,
    axis x line=none,
    y label style={at={(axis description cs:1.1,0.5)},anchor=south, yshift=-5em},
]S

\addplot[green!60!black, thick, dashed, mark=triangle*, smooth] coordinates {
    (0,10)
    (1,49.63)
    (2,65.14)
    (3,70.83)
    (4,73.04)
    (5,74.28)
};
\addlegendentry{Accuracy}

\end{axis}

\end{tikzpicture} } \vspace{-5pt}
\caption{Reconstruction quality (SSIM) and model accuracy across training epochs. Batch Size 8}
\label{epoch_fig}  \vspace{-10pt}
\end{figure}

\textbf{Robustness to batch size.} Figure \ref{batch_fig} demonstrates the effect of varying client batch size on reconstruction MSE. All attacks suffer performance degradation with larger batches due to the aggregation of more diverse gradients, which dilutes per-sample signal. However, our method consistently yields lower MSE across all batch sizes, demonstrating that it is less sensitive to this form of information obfuscation. The relative gap between our method and competing approaches widens with larger batches, suggesting that our generative prior mitigates the signal dilution problem more effectively. This highlights the robustness of our approach under realistic federated learning configurations. We note that the GGL attack tends to align more closely with the ground truth on single-image batches, rather than producing fully realistic but generative reconstructions.

\textbf{Resilience to defenses.} Table \ref{tab:deep_leakage_def_metrics} evaluates attack performance under standard defenses (Gaussian noise, clipping, sparsification and Precode). Across Gaussian noise injection, weight clipping, and weight sparsification, our method consistently achieves the best SSIM and lowest LPIPS/MSE values, often by a clear margin. Notably, under sparsification, a defense that simultaneously offers communication efficiency gains our method preserves a high SSIM of 0.554 compared to 0.512 for SME and substantially lower values for other baselines. This indicates that our generative prior is robust not only to stochastic perturbations but also to structured distortions of the gradient signal. While defenses do degrade all methods to some extent, our attack remains the most effective, underscoring its resilience in realistic FL environments. Even under Precode, which introduces a stronger bottleneck, our approach maintains the highest reconstruction quality, though the margin over competing attacks is smaller compared to other defenses.

\begin{figure}
    \centering
    \includegraphics[width=1\linewidth]{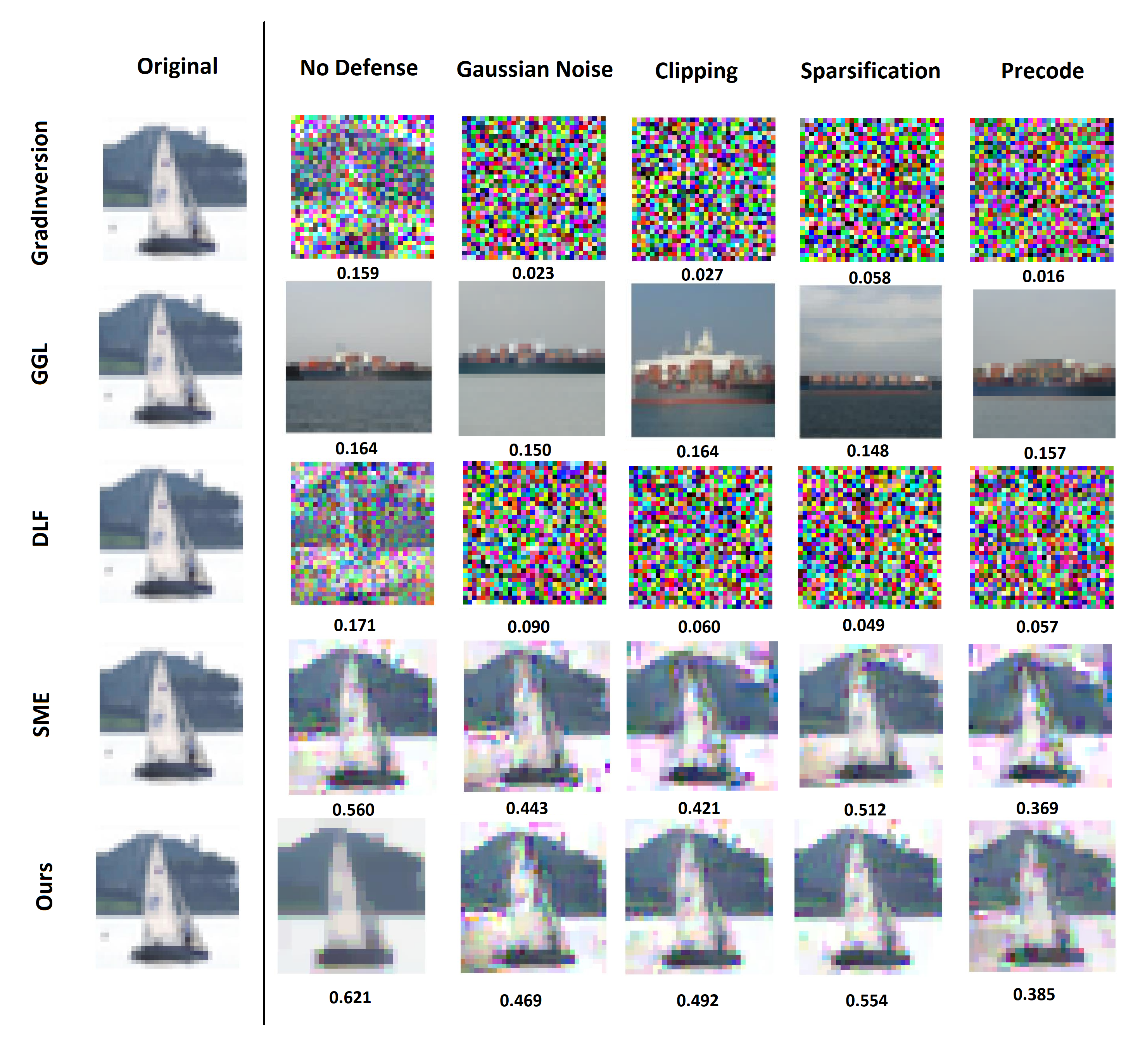}
    \caption{Qualitative reconstruction comparisons across multiple attack and defense methods. SSIM scores from Table \ref{tab:deep_leakage_metrics} and \ref{tab:deep_leakage_def_metrics} included below.}
    \vspace{-15pt}
    \label{fig:visual}
\end{figure}

\textbf{Effect of Defense Parameter.} As shown in Table \ref{tab:noise_vs_ssim}, increasing the standard deviation of injected Gaussian noise reduces attack success but also severely degrades model accuracy. For example, while small noise levels ($\sigma$ = $1 \times 10^{-3}$) only moderately impact reconstruction quality (SSIM $\approx$ 0.469), larger perturbations ($\sigma$ = $2 \times 10^{-3}$) collapse both accuracy and reconstruction. This illustrates the inherent trade-off between preserving model utility and providing meaningful privacy guarantees, motivating the need for stronger defenses beyond DP based techniques.

\textbf{Performance across epochs.} Figure \ref{epoch_fig} shows the evolution of reconstruction quality (SSIM) and model accuracy over global training epochs. Our method consistently outperforms existing attacks, particularly in later training stages where FedAvg produces smaller, more homogeneous updates that carry less distinct information about individual data samples making reconstruction more challenging. This indicates that our generative prior is particularly effective at extracting meaningful image content even from noisy, unstable updates. In contrast, baseline methods exhibit stagnation or high variance in later stages, suggesting less stable optimization.

\textbf{Model architecture comparisons.} Table \ref{tab:deep_leakage_models} shows that our attack succeeds across a variety of architectures. Results are reported after 2 epochs of global training on CIFAR-10, but because different models train at different rates and attacks are sensitive to training state (Figure \ref{epoch_fig}), direct comparisons are difficult. Nevertheless we can see that higher quality reconstructions are obtained from simpler feed forward models (MLP and ConvNet). VGG proves the hardest network to attack, likely because it's depth and lack of residual layers lead to attenuated gradients.

\textbf{Qualitative results.} Figure \ref{fig:visual} provides visual comparisons across methods. Our attack produces sharper images with more coherent object boundaries and semantically accurate details, whereas baseline methods frequently yield blurry textures, distorted structures, or artifacts inconsistent with the target class. The qualitative improvements corroborate the quantitative metrics, highlighting that our approach not only minimizes numerical error but also recovers reconstructions that are visually more recognizable and useful to an adversary. While GGL reconstructions appear more coherent, this coherence stems from leveraging a GAN trained on an external dataset. The attack recovers a class-consistent image, but it is not a faithful reconstruction of the original private data. This is why in Table \ref{tab:deep_leakage_metrics} GGL has a very competitive TV score but doesn't compete in any similarity metric.
\vspace{-20pt}
\section{Conclusion}
\label{conc}
\vspace{-6pt}
We presented a new deep leakage attack that integrates a generative flow matching denoiser into the reconstruction process. Unlike prior methods, our approach directly regularizes the inversion by constraining reconstructions toward the data distribution modeled by a flow matching model trained on external data. This design enables the attack to exploit both the information contained in client updates and the structural knowledge embedded in the generative prior, producing reconstructions that are sharper, more realistic, and semantically aligned with the target class.

Our evaluation on CIFAR-10 and Tiny-ImageNet shows that this distribution-driven regularization significantly improves reconstruction quality over state-of-the-art baselines. A key finding of this work is that the generative prior does not need to be trained on the same dataset as the federated learning task (which isn't feasible in real attack scenarios). Even when the flow matching model is trained on unrelated data, it provides sufficient distributional guidance to substantially enhance reconstructions on novel datasets. The method consistently achieves higher PSNR and SSIM, lower perceptual errors (LPIPS, FMSE), and maintains strong performance across varying batch sizes and training epochs. Moreover, the attack remains robust against common defenses such as noise injection, clipping, and sparsification, underscoring the resilience of distributionally guided reconstructions.

These findings highlight the heightened privacy risks posed by adversaries equipped with generative priors. Even when defenses are deployed, realistic data distributions can be exploited to recover private information at high fidelity. This work emphasizes the urgent need for defenses that explicitly account for distribution-aware attacks, and points toward a broader research agenda on privacy risks in FL under the lens of powerful generative models.



\bibliography{example_paper}
\bibliographystyle{icml2026}

\end{document}